\documentclass[final]{article} 
\usepackage{colm2024_conference}

\usepackage{microtype}
\usepackage{hyperref}
\usepackage{url}
\usepackage{booktabs}

\usepackage{multirow, amssymb, amsmath, wrapfig,lipsum}
\usepackage{pifont}
\newcommand{\cmark}{\ding{51}}%
\newcommand{\xmark}{\ding{55}}%

\usepackage{array}
\newcolumntype{H}{>{\setbox0=\hbox\bgroup}c<{\egroup}@{}}
\newcolumntype{x}[1]{>{\centering\arraybackslash}p{#1}}

\usepackage{graphicx, caption,enumitem}

\title{Context-Aware Clustering using Large Language Models}

\author{%
Sindhu Tipirneni$^{1}\thanks{Corresponding author: sindhut@vt.edu}~~$  Ravinarayana Adkathimar$^{2}~~$ Nurendra Choudhary$^2$  
\\ \textbf{Gaurush Hiranandani}$^2~~$  \textbf{Rana Ali Amjad}$^2~~$  \textbf{Vassilis N. Ioannidis}$^2~~$  \textbf{Changhe Yuan}$^2~~$ \\ \textbf{Chandan K. Reddy}$^{1,2}$ \\
$^1$Department of Computer Science, Virginia Tech, Arlington, VA, USA\\
$^2$Amazon, Palo Alto, CA, USA
}

\colmfinalcopy 
\begin{document}

\maketitle

\vspace{-3ex}
\begin{abstract}

Despite the remarkable success of Large Language Models (LLMs) in text understanding and generation, their potential for text clustering tasks remains underexplored. We observed that powerful closed-source LLMs provide good quality clusterings of entity sets but are not scalable due to the massive compute power required and the associated costs. Thus, we propose CACTUS (Context-Aware ClusTering with aUgmented triplet losS), a systematic approach that leverages open-source LLMs for efficient and effective supervised clustering of entity subsets, particularly focusing on text-based entities. 
Existing text clustering methods fail to effectively capture the context provided by the entity subset. 
Moreover, though there are several language modeling based approaches for clustering, very few are designed for the task of supervised clustering.
This paper introduces a novel approach towards clustering entity subsets using LLMs by capturing context via a scalable inter-entity attention mechanism. 
We propose a novel augmented triplet loss function tailored for supervised clustering, which addresses the inherent challenges of directly applying the triplet loss to this problem. 
Furthermore, we introduce a self-supervised clustering task based on text augmentation techniques to improve the generalization of our model. 
For evaluation, we collect ground truth clusterings from a closed-source LLM and transfer this knowledge to an open-source LLM under the supervised clustering framework, allowing a faster and cheaper open-source model to perform the same task.
Experiments on various e-commerce query and product clustering datasets demonstrate that our proposed approach significantly outperforms existing unsupervised and supervised baselines under various external clustering evaluation metrics.

\end{abstract}

\vspace{-2ex}
\section{Introduction}
\begin{wrapfigure}{r}{0.5\textwidth}
  \begin{center}
    \includegraphics[width=0.48\textwidth,trim=0 0ex 0 0ex]{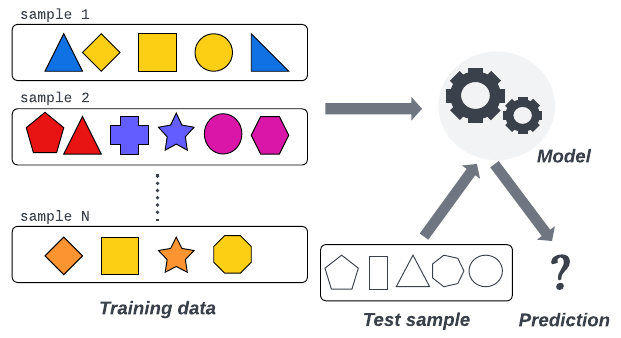}
  \end{center}
    \caption{Illustration of the supervised clustering problem: Each training sample contains a subset of entities along with the corresponding ground truth clustering. Given a test sample, which is an unseen entity subset, the goal is to cluster the entities in the test sample. In a sample, color denotes a cluster, and shape denotes an entity.}
    \label{fig:sup_clus}
\end{wrapfigure}

Large Language Models (LLMs) have demonstrated human-level performance in text understanding and generation, but their application to text clustering tasks is underexplored. We observed that powerful closed-source LLMs (such as GPT-4 \citep{achiam2023gpt} and Claude \citep{claude}), known for their instruction-following abilities, can provide high-quality clusterings through prompting. However, these models become unaffordable when clustering a large number of sets, due to their high costs. To overcome this limitation, we aim to develop a scalable model based on an open-source LLM that can efficiently and effectively perform the clustering task. We study this problem of transferring the knowledge of clustering task from a powerful closed-source LLM ($\text{LLM}_\text{c}$) to a scalable open-source LLM ($\text{LLM}_\text{o}$) under the framework of supervised clustering, where the goal is to learn to cluster unseen entity subsets, given training data comprising several examples of entity subsets with complete clusterings\footnote{Complete clustering of a set refers to a clustering in which every entity in the set is assigned to a cluster.} (See Figure \ref{fig:sup_clus}).

In this work, we focus particularly on entities described by text.
This problem has applications in various domains including e-commerce, news clustering, and email management, among others \citep{finley2005supervised,finley2008supervised, haider2007supervised}.
However, deep learning approaches for solving the supervised clustering  problem remain largely unexplored. Existing methods overlook the specific context provided by an entity subset and often rely on the latent structural loss function \citep{fernandes2012latent} which involves the sequential computation of maximum spanning forests.  In our work, 
we propose an $\text{LLM}_\text{o}$-based solution called CACTUS (Context-Aware ClusTering with aUgmented triplet losS) that captures contextual information, introduces an enhanced loss function, and incorporates a self-supervised clustering task.

The context of an entity subset refers to the unique circumstances that relate the specific entities occurring in the subset.
For example, consider the task of clustering a user's monthly purchases. A purchase of `magnetic tape' could signify various intentions, such as for a science project or picture mounting. Examining the user's other purchases could provide the necessary context to help us determine the use case and place the entity in the appropriate cluster. However, most existing text clustering methods obtain a single embedding for each entity using a language model \citep{ahmed2022short,Barnabo2023}, thus ignoring the context. In contrast, our model computes entity embeddings that are dependent on the context or entity subset, which allows the model to identify entities with shared themes within the subset. The proposed method takes the entire entity subset as input to the LLM and captures inter-entity interactions using a scalable attention mechanism, as traditional full attention over all entities in a subset can become computationally expensive as subsets grow large. Specifically, in each Transformer layer,
for each entity, we compute a single representative embedding that participates in inter-entity attention.

Previous methods for supervised clustering applied the latent structural loss to pairwise entity features that are either hand-crafted or obtained from a neural network. 
While the latent structural loss involves sequential computations of spanning forests, the triplet loss can be parallelized (processing all triplets in a clustering in parallel using more memory) but faces the challenge of different triplets potentially having non-overlapping margin positions (see section \ref{sec:loss}). 
To address this issue, we augment the complete graph of entities with a neutral entity, which is connected to all other entities by a learnable similarity score that provides a reference for all margin locations. 
Additionally, to further improve supervised clustering performance, especially in the case of limited availability of ground truth clusterings, we introduce a novel self-supervised clustering task. 
This task involves randomly sampling seed entities and constructing clusters with different transformations of each seed. 
This idea is inspired by text data augmentation techniques \citep{shorten2021text} used in NLP tasks, but we formulate it, for the first time, as a self-supervised clustering task that aligns better with our finetuning phase. 

To summarize, the main contributions of our work are as follows:
\begin{itemize}[leftmargin=*]
    \item We propose a novel approach for supervised clustering of entity subsets using context-aware entity embeddings from  $\text{LLM}_\text{o}$ with a scalable inter-entity attention mechanism. 
    \item  We identify a problem with directly applying triplet loss to supervised clustering when different triplets can potentially have non-overlapping margin locations. To address this, we design a novel augmented triplet loss function.
    \item  We also design a self-supervised clustering task to improve $\text{LLM}_{o}$'s finetuning performance, especially when only a limited number of ground truth clusterings are available.
    \item Our experiments demonstrate that the proposed method, CACTUS, outperforms previous unsupervised and supervised clustering baselines on real-world e-commerce query and product clustering datasets.
    We also conduct ablation studies to show the effectiveness of each of the proposed components.
\end{itemize}

\section{Related work}

\subsection{Traditional methods for supervised text clustering}
The supervised clustering problem can be formulated as a binary pairwise classification task of predicting if a pair of entities belong to the same cluster. But this approach suffers from the drawback that the pairs are assumed to be i.i.d. \citep{finley2005supervised}. Thus, structured prediction approaches have been explored as solutions to this problem. Traditional methods used hand-engineered pairwise features as inputs, where each pair of entities is described by a vector. 
Methods such as structural SVM \citep{tsochantaridis2004support, finley2005supervised} and structured perceptron \citep{collins2002discriminative} have been applied to this problem, where a parameterized scoring function is learned such that it assigns higher scores to correct clusterings in the training data. 
The scoring function depends on the pairwise features and the predicted clustering, and is formulated using correlation clustering \citep{bansal2004correlation} or k-means \citep{finley2008supervised} frameworks. Observing that many within-cluster entity pairs have weak signals, \citet{yu2009learning, fernandes2012latent, haponchyk2018supervised} introduce maximum spanning forests over complete graphs of entities as latent structures in the scoring function. The inference stage involves finding a clustering with the highest score for a given entity subset.

\subsection{Language models for text clustering}
Despite the widespread use of Language Models (LMs) across diverse domains and applications, their application to `supervised' clustering remains limited. 
\citet{haponchyk2021supervised} and \cite{Barnabo2023} utilize encoder-only LMs to obtain pairwise and individual entity representations, respectively, and finetune the LMs using latent structural loss. The former is not a scalable approach as each entity pair is passed separately through a conventional Transformer model. 
In contrast to these existing methods, we propose a novel approach that passes the entire entity set to a language model, and efficiently models inter-entity interactions within the Transformer layers, thereby improving clustering performance by capturing the unique context given by an entity subset. Furthermore, we depart from the latent structural loss (used in these existing works) that involves the sequential step of computing maximum spanning forests and employ an augmented triplet loss function that can be more easily parallelized and also achieves better performance.

It is worth noting that LMs have been widely applied to slightly different but more prevalent problems of unsupervised \citep{bertopic,zhang2021supporting,zhang2021short,meng2022topic} and semi-supervised clustering \citep{zhang2021discovering, lin2020discovering,zhang2022new,an2023generalized}. 
These tasks involve clustering of a single large entity set, with some pairwise constraints provided for semi-supervised clustering. Some recent works \cite{viswanathan2023large, zhang2023clusterllm, nakshatri2023using} take advantage of the latest advances in LLMs by using them as oracles to make key decisions during the clustering process. However, these approaches are not suitable for our problem of clustering several entity subsets, as they require a new optimization problem for every new entity subset. 
Different from these LLM-based methods, our approach involves prompting $\text{LLM}_\text{c}$ to gather complete clusterings of several small entity subsets, which are subsequently used to fine-tune a scalable $\text{LLM}_\text{o}$ that is adapted to capture the underlying context efficiently.

\section{Proposed Method}
\begin{figure}
    \centering
    \includegraphics[scale=0.45]{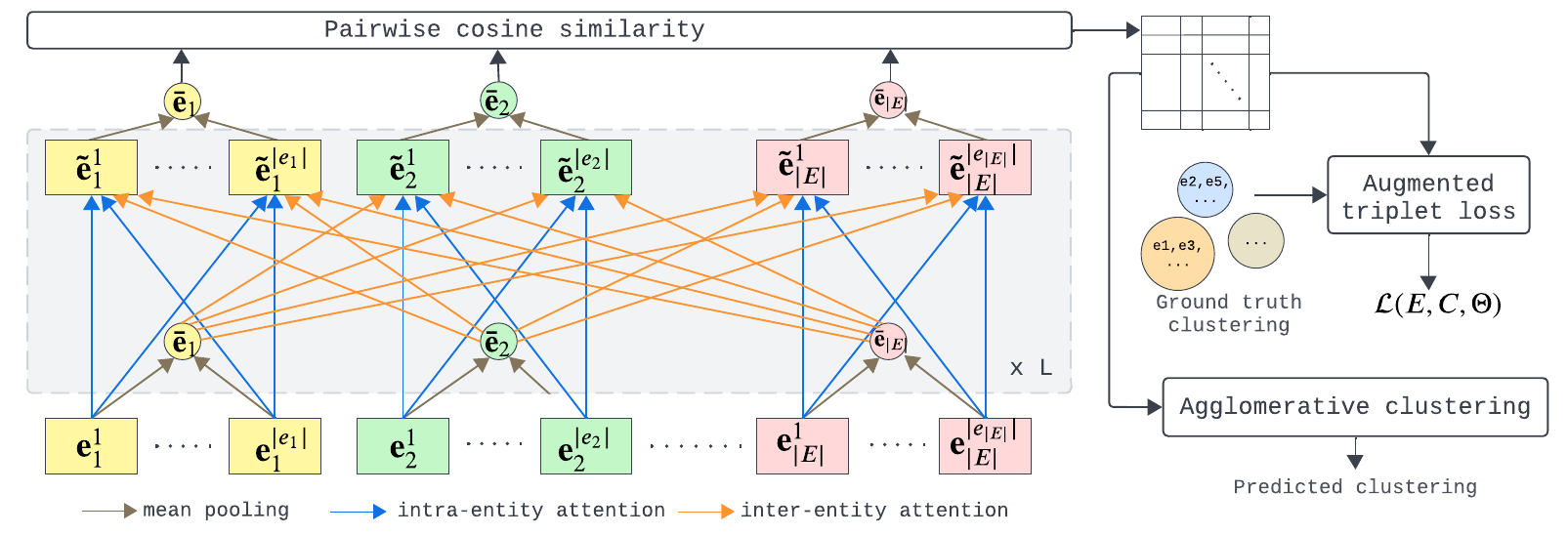}
\caption{Overview of CACTUS: The entities in the input subset are tokenized and passed through $\text{LLM}_\text{o}$, where the self-attention layers are modified with scalable inter-entity attention (SIA) to obtain context-aware entity embeddings. Pairwise cosine similarities are used for computing loss and predicted clusterings.}
    \label{fig:sia}
\end{figure}

This section provides a detailed description of the supervised clustering problem and our proposed method. Our approach involves finetuning an open-source pretrained Transformer encoder model, denoted by $\text{LLM}_\text{o}$, for the task of context-aware clustering in a supervised manner. Here, `context' refers to the subset in which an entity occurs, which influences the entity's interpretation. 
To capture context-awareness efficiently, we modify the self-attention layers of $\text{LLM}_\text{o}$ to implement a scalable inter-entity attention mechanism, which is described in section \ref{sec:caee}. We identify limitations of directly applying the triplet loss to supervised clustering and propose an augmented triplet loss function as a solution in section \ref{sec:loss}. We further pretrain $\text{LLM}_\text{o}$ on a dataset-specific self-supervised clustering task before the finetuning phase, which is described in Appendix \ref{sec:pretraining} due to space constraints. During inference, given an entity subset, we extract context-aware entity embeddings from the finetuned model, compute pairwise similarities, and feed them to an agglomerative clustering algorithm to obtain the predicted clustering. We refer to the overall method as CACTUS (Context-Aware ClusTering with aUgmented triplet losS). Figure \ref{fig:sia} provides an overview of the proposed approach.

\subsection{Preliminaries}

Let $\mathcal{E}$ be the universal set of entities in a dataset.
For an entity subset $E\subseteq\mathcal{E}$, a clustering $\mathcal{C}=(C,f)$ contains a set of clusters $C$ and an entity-to-cluster assignment function $f:E\twoheadrightarrow C$\footnote{$\twoheadrightarrow$ denotes a surjective function.}.
We say that two clusterings, $\mathcal{C}=(C,f)$ and $\mathcal{C}'=(C',f')$, over the same enitity subset $E$ are equivalent if they induce the same partitioning of items i.e., if the pairwise co-cluster relationships are preserved. Formally, the clusterings $\mathcal{C}$ and $\mathcal{C}'$ are equivalent if and only if $\forall e_1,e_2 \in E$, we have
\begin{align}
    f(e_1) = f(e_2) \iff f'(e_1) = f'(e_2).
\end{align}
A labeled clustering dataset $\mathcal{D} = \{(E_1,\mathcal{C}_1), ..., (E_{|\mathcal{D}|},\mathcal{C}_{|\mathcal{D}|})\}$ contains $|\mathcal{D}|$ samples where each sample contains an entity subset $E_k\subseteq \mathcal{E}$ and the corresponding ground truth clustering $\mathcal{C}_k$. 
We describe the process of collecting cluster assignments from $\text{LLM}_\text{c}$ in Appendix \ref{sec:ground}. These clusterings serve as ground truth in the dataset, which is partitioned into training, validation, and test splits.
Given an entity subset $E_k$, our goal is to predict a clustering that is equivalent to the ground truth clustering $\mathcal{C}_k$. We use $\texttt{text($e$)}$ to denote the textual description of entity $e$. 


\subsection{Context-awareness using Scalable Inter-entity Attention (SIA)} \label{sec:caee}
Here, we describe the architecture of $\text{LLM}_\text{o}$, which is a Transformer encoder model that is finetuned on the supervised clustering task using ground truth clusterings from $\text{LLM}_\text{c}$. 
A common approach for text clustering involves obtaining a single embedding vector separately for each entity using a language model and defining a similarity or distance function in the embedding space, which is used in a clustering algorithm. 
We refer to this approach as NIA (No Inter-entity Attention) because there is no interaction between different entities in the embedding module. 
To capture context, i.e., to model entity embeddings that depend on the entity subset they occur in, we can also pass the entire subset in the input sequence and pool each entity's token embeddings. 
We refer to this approach as FIA (Full Inter-entity Attention), because all the token pairs among different entities are considered in the attention matrix. This is not very practical, especially, when entity descriptions are long and as the subsets grow large. 
So, we design a scalable inter-entity attention (SIA) mechanism that computes one representative embedding per entity which is used for inter-entity interactions. 
Though there are scalable attention methods for handling long sequences in Transformers \citep{longformer,reformer,etc}, this is the first work to explore scalable attention in the context of clustering. 
The proposed SIA approach is described in detail below. We use the encoder of Flan-T5-base \citep{chung2022scaling} as the underlying model and modify its attention layers for SIA.

Let $E=\{e_1,...,e_{|E|}\}$ be an entity subset, where tokens of entity $e_i$ are denoted as $tokenize(text(e_i)) = (e_i^1,...,e_i^{|e_i|})$. 
A Transformer-based LM gathers initial token embeddings and iteratively updates them using stacked Multi-Head Attention (MHA) and Feed Forward Network (FFN) layers. 
The Multi-Head Attention (MHA) layer traditionally computes all token-token pairwise attention scores, making it computationally intensive for long inputs. 
In SIA mechanism, we propose modifications to the MHA layer to make it more scalable for our clustering task. We split the attention computation into intra-entity and inter-entity components and make the latter more efficient by using pooled entity representations. 
Let $\mathbf{e_i^j} \in \mathbb{R}^d$ denote the embedding of token $e_i^j$ ($j^{th}$ token of $i^{th}$ entity) in the input to an MHA layer, and $\mathbf{\bar{e}_i}=\frac{1}{|e_i|}\sum_k\mathbf{e_i^k}$ denote the mean-pooled representation of entity $e_i$. 
The MHA layer transforms the embedding $\mathbf{e_i^j}$ to $\mathbf{\tilde{e}_i^j}\in \mathbf{R}^{d}$ as follows. For simplicity, we show the computations for a single attention head and skip the projection layer at the end of MHA.
\begin{align}
\mathbf{\tilde{e}_i^j} &= \underbrace{\sum_{k=1}^{|e_i|} \alpha_{intra} (e_i^j,e_i^k)  \mathbf{W^V} \mathbf{e_i^k}}_\text{intra-entity attention} + \underbrace{\sum_{\substack{m=1\\m\neq i}}^{|E|} \alpha_{inter}(e_i^j, e_m) \mathbf{W^V} \mathbf{\bar{e}_m}
}_\text{inter-entity attention} \label{eq:main_sia}\\
\alpha_{intra(inter)}(e_i^j,.) &= \frac{exp(Att_{intra(inter)}(e_i^j,.))}{\sum_{k=1}^{|e_i|} exp(Att_{intra}(e_i^j,e_i^k)) + \sum_{\substack{m=1\\m\neq i}}^{|E|} exp(Att_{inter}(e_i^j,e_m))} \label{eq:alpha} \\
Att_{intra}(e_i^j,e_i^k) &= (\mathbf{W^Qe_i^j})^T(\mathbf{W^Ke_i^k}) + \phi(k-i) \label{eq:intra_att}\\
Att_{inter}(e_i^j, e_m) &= (\mathbf{W^Qe_i^j})^T (\mathbf{W^K}\mathbf{\bar{e}_m}) \label{eq:inter_att}
\end{align}

where $\mathbf{W^Q},\;\mathbf{W^K},\;\mathbf{W^V}\;\in \mathbb{R}^{d\times d}$ are the query, key, and value projection matrices, respectively. 
Eq. \eqref{eq:main_sia} shows that a token within one entity attends to aggregated representations of other entities rather than individual tokens within those entities. The traditional softmax computation is altered in \eqref{eq:alpha} to separate the intra and inter-entity terms. 
The intra-entity attention \eqref{eq:intra_att} includes a relative positional encoding term, denoted by $\phi(.)$, while the inter-entity attention \eqref{eq:inter_att} does not. This is because the order of tokens within an entity is relevant while the order of entities in a subset is irrelevant. The token embeddings from the last Transformer layer are mean-pooled entity-wise to obtain the context-aware entity embeddings.

\textbf{Complexity: } Considering a subset of $N$ entities where each entity contains $L$ tokens, and a fixed embedding dimension $d$, the computational complexity of self-attention in the NIA embedding method is $O(NL^2)$ because there are $NL$ tokens in the entity subset, and each token only attends to the $L$ tokens within the same entity. In contrast, using the FIA approach increases the complexity to $O(N^2L^2)$ as each token attends to all $NL$ tokens from all entities. 
SIA provides a compromise between these two methods; it has $O(NL(L+N))$ complexity because each token attends to the $L$ tokens within the same entity and to $N-1$ representative entity embeddings.

\subsection{Augmented triplet loss} \label{sec:loss}
After obtaining context-aware entity embeddings, we compute cosine similarity between all entity pairs in a subset:
\begin{align}
    \operatorname{sim}(e_i, e_k) = \frac{\mathbf{\bar{e}_i^\top \bar{e}_k}}{\|\mathbf{\bar{e}_i}\|\|\mathbf{\bar{e}_k}\|}
\end{align}
The similarities are used to obtain predicted clusterings using the average-link agglomerative clustering algorithm. 
For the loss function, using these pairwise similarities as edge weights, we can construct a fully connected graph where each entity is a node. 
Previous methods for supervised clustering employed structural loss, which uses a scoring function based on a maximum spanning forest of the fully connected graph. This uses Kruskal's MST algorithm, which sequentially adds edges to the spanning forest and leads to slower loss computation. In contrast, the triplet loss \citep{schroff2015facenet}, which was shown to be a competitive baseline in \cite{Barnabo2023}, can be easily parallelized as each triplet can be processed independently of the others. For each entity in the input subset, the triplet loss considers other entities within the same cluster as positives and the remaining entities as negatives. For an entity subset $E$ with ground truth clustering $\mathcal{C}=(C,f)$, the triplet loss is given by 
\begin{align}
    \mathcal{L}^{triplet}(E,\mathcal{C},\Theta) = \frac{1}{|T(\mathcal{C})|} \sum_{(e,e_p,e_n)\in T(\mathcal{C})} (\gamma - \operatorname{sim}(e,e_p) + \operatorname{sim}(e,e_n))^+\label{eq:triplet}
\end{align}
where $\Theta$ are the parameters of the context-aware entity embedding module, $\gamma$ is the margin which is a hyperparameter, and $T(\mathcal{C})=\{(e,e_p,e_n):e,e_p,e_n\in E; e\neq e_p; f(e)=f(e_p)\neq f(e_n)\}$ is the set of triplets.

\begin{wrapfigure}{r}{0.35\textwidth}
  \begin{center}
\includegraphics[scale=0.4,trim=0 0 0 5ex]{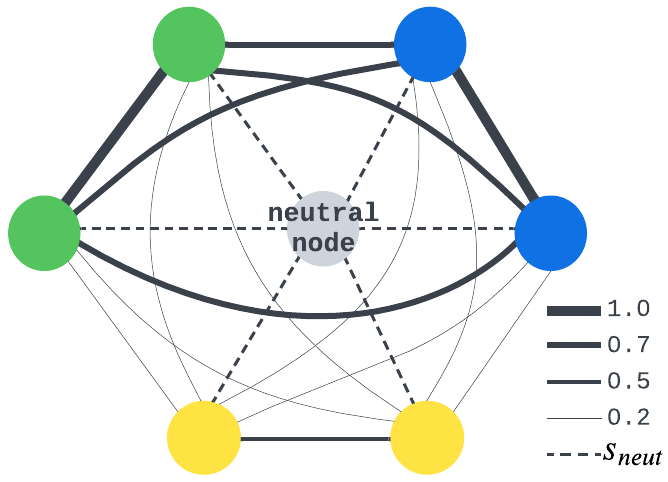}
  \end{center}
\caption{Example of an entity subset with 3 clusters containing 2 entities each. There exists an intra-cluster (yellow) edge with similarity less than some inter-cluster (green-blue) edges. For margin=0.3, the triplet loss (eq. \ref{eq:triplet}) is at its minimum while the proposed augmented triplet loss (eq. \ref{eq:aug_triplet}) is not.}
\label{fig:loss}
\end{wrapfigure}
The triplet loss formulation presents a challenge due to potential non-overlapping margin locations across different triplets. 
Margin location refers to the range between similarities from anchor entity ($e$) to positive ($e_p$) and negative ($e_n$) entities within a triplet. 
For example, in Figure \ref{fig:loss} with three clusters containing two entities each, the pairwise similarities shown result in the minimum value for triplet loss. However, there exist inter-cluster edges with higher similarity than an intra-cluster edge, which results in `green' and `blue' clusters being merged by the agglomerative clustering algorithm before the `yellow' cluster is formed. This phenomenon can also occur for intra and inter-cluster edges in different entity subsets, which makes it difficult to choose a global threshold for agglomerative clustering during inference. To avoid such problems, we augment the complete graph with a neutral node that is connected to all other entities via a learnable neutral edge similarity $s_{neu}$. The neutral node is incorporated into the augmented triplet loss to encourage intra and inter-cluster edge similarities to lie on opposite sides of $s_{neu}$. The new loss function is given by 
\begin{align}
    \mathcal{L}^{aug\text{-}triplet}&(E,\mathcal{C},\Theta) = \tfrac{1}{|T(\mathcal{C})|+|P^{intra}(\mathcal{C})|+|P^{inter}(\mathcal{C})|}\bigg\{ \sum_{(e,e_p,e_n)\in T(\mathcal{C})} (\gamma - \operatorname{sim}(e,e_p) + \operatorname{sim}(e,e_n))^+ \nonumber\\
    + & \sum_{(e,e_p)\in P^{intra}(\mathcal{C})} (\frac{\gamma}{2}-\operatorname{sim}(e,e_p)+s_{neu})^+ 
    + \sum_{(e,e_n)\in P^{inter}(\mathcal{C})} (\frac{\gamma}{2}-s_{neu}+\operatorname{sim}(e,e_n))^+\bigg\} \label{eq:aug_triplet}
\end{align}
where $P^{intra}(\mathcal{C})=\{(e,e_p):(e,e_p,.)\in T(\mathcal{C})\}$ is the set of entity pairs in the same cluster and $P^{inter}(\mathcal{C})=\{(e,e_n):(e,.,e_n)\in T(\mathcal{C})\}$ is the set of entity pairs in different clusters. 
The newly added loss terms encourage the intra-cluster (inter-cluster) pairwise similarities to be $\frac{\gamma}{2}$ higher (lower) than the neutral edge similarity. Thus, the neutral edge softly constraints the margin location for all triplets.

\section{Experiments} \label{sec:exp}

In this section, we describe the datasets used for our experiments and compare the proposed method to existing unsupervised and supervised clustering baselines using external clustering evaluation metrics. Additionally, we conduct ablation studies to analyze the effectiveness of the different components of our method. Finally, we present a qualitative study to illustrate how context-awareness improves clustering performance.

\begin{table}[]
    \centering
    \small
        \caption{Dataset statistics. (* Since the Gifts dataset is proprietary, we provide approximate numbers for the statistics reported.)}
    \label{tab:data_stats}
    
    \begin{tabular}{lccccc}
    \toprule
    & Gifts* & Arts & Games & Instruments & Office \\
    \midrule
    No. of entities & $\sim$365K & 22,595 & 16,746 & 10,522 & 27,532 \\
    No. of entity sets & $\sim$42K & 55,629 & 54,995 & 27,420 & 100,775 \\
    Avg. size of entity set & $\sim$46 & 5.4 & 5.7 & 5.6 & 5.0 \\
    Avg. no. of clusters per entity set & $\sim$6 & 2.6 & 2.8 & 2.8 & 2.7 \\
    Avg. no. of entities per cluster & $\sim$8 & 2.1 & 2.1 & 2.0 & 1.9 \\
    Avg. no. of words per entity & $\sim$3 & 11.6 & 6.9 & 10.5 & 13.9  \\
    \bottomrule
    \end{tabular}

\end{table}

\subsection{Experimental setup}
We compile five datasets for our experiments, including four from Amazon product reviews \citep{ni2019justifying} and one proprietary dataset called Gifts. The Amazon datasets including Arts, Games, Instruments, and Office, consist of sequences of products reviewed by users, with each user's product sequence treated as one entity subset. We use preprocessed datasets from \cite{recformer}, considering product titles as textual descriptions of entities. The Gifts dataset contains search queries related to `gifts' from an e-commerce platform. Each entity subset contains potential next queries for a particular source query. 
Dataset statistics are summarized in Table \ref{tab:data_stats}. On average, the Amazon datasets contain 5 to 6 entities per entity subset, while Gifts contains approximately 46 entities. 
In each dataset, we randomly sample 3K entity sets for test split and 1K sets for validation split and use the remaining for training. 
For all datasets, we use a proprietary $\text{LLM}_\text{c}$ to collect ground truth clusterings. 
We run self-supervised pretraining for the Amazon datasets but not for Gifts, as the queries in Gifts are very short, making it difficult to obtain multiple transformations of a query. 
We evaluate the predicted clusterings from $\text{LLM}_\text{o}$ by comparing them to ground truth clusterings. 
Thus, we use the following extrinsic clustering evaluation metrics: Rand Index (RI), Adjusted Rand Index (ARI), Normalized Mutual Information (NMI), Adjusted Mutual Information (AMI), and F1-score \cite{haponchyk2018supervised}. 


\subsection{Comparison with baselines}
As unsupervised clustering baselines, we employ the K-Means, Spectral, and Agglomerative clustering algorithms. The entity embeddings for unsupervised baselines are obtained from the pretrained Flan-T5-base encoder. For K-Means and Spectral clustering, we determine the number of clusters for each entity set using either the silhouette method or the average number from the training set based on validation metrics. 
For agglomerative clustering, we use cosine similarity with average linkage and determine the threshold based on the validation set. 
Given the scarcity of existing supervised clustering baselines, we incorporate only one such method from \citet{Barnabo2023} (SCL). 
NSC \citep{haponchyk2021supervised} was not included as it demands substantial GPU memory and often leads to OOM errors. 
For a fair comparison, we employ FlanT5-base encoder as the LLM for all baselines and the results are shown in Table \ref{tab:baselines}. 
CACTUS significantly outperforms all the unsupervised and supervised baselines. 
Compared to SCL, CACTUS improves the AMI and ARI metrics by 12.3\%-26.8\% and 15.3\%-28.2\%, respectively. 
Among the unsupervised methods, agglomerative clustering yields the best result in most cases.

\begin{table}
\setlength{\tabcolsep}{3pt}
\begin{minipage}{0.55\linewidth}
    \centering
    \small
    \caption{Comparison of the proposed method to previous unsupervised and supervised clustering baselines. The first three are unsupervised methods and the last two are supervised clustering methods. (Agglo. stands for agglomerative clustering. *For the proprietary Gifts dataset, we report improvements against K-Means.)}
    \label{tab:baselines}
    \begin{tabular}{lp{36pt}x{25pt}x{25pt}x{25pt}x{25pt}x{25pt}}
\toprule
 & Model & NMI & AMI & RI & ARI & F1 \\

\midrule
\parbox[t]{2mm}{\multirow{5}{*}{\rotatebox[origin=c]{90}{Gifts*}}}

& K-Means & +0.000 & +0.000 & +0.000 & +0.000 & +0.000\\
& Spectral & +0.020 & +0.024 & -0.002 & +0.006 & +0.014 \\
&  Agglo. & +0.047 & +0.009 & -0.019 & +0.011 & +0.027\\
& SCL & \underline{+0.167} & \underline{+0.196} & \underline{+0.065} & \underline{+0.195} & \underline{+0.114} \\
& CACTUS & \textbf{+0.207} & \textbf{+0.260} & \textbf{+0.098} & \textbf{+0.263} & \textbf{+0.144} \\

\midrule
\parbox[t]{2mm}{\multirow{5}{*}{\rotatebox[origin=c]{90}{Arts}}}
& K-Means & 0.660 & 0.167 & 0.690 & 0.250 & 0.766 \\
& Spectral & 0.642 & 0.192 & 0.683 & 0.272 & 0.790 \\
& Agglo. & 0.692 & 0.219 & 0.707 & 0.290 & 0.781  \\
& SCL &  \underline{0.725} &  \underline{0.371} &  \underline{0.751} &  \underline{0.435} &  \underline{0.833} \\
& CACTUS & \textbf{0.764} & \textbf{0.461} & \textbf{0.795} & \textbf{0.540} & \textbf{0.868} \\

\midrule
\parbox[t]{2mm}{\multirow{5}{*}{\rotatebox[origin=c]{90}{Games}}}
& K-Means & 0.681 & 0.213 & 0.712 & 0.247 & 0.767 \\
& Spectral & 0.688 & 0.230 & 0.718 & 0.263 & 0.771 \\
&  Agglo. & 0.640 & 0.268 & 0.691 & 0.291 &0.799 \\
& SCL &  \underline{0.718} &  \underline{0.442} &  \underline{0.763} &  \underline{0.462} &  \underline{0.849} \\
& CACTUS & \textbf{0.777} & \textbf{0.540} & \textbf{0.813} & \textbf{0.565} & \textbf{0.876} \\

\midrule
\parbox[t]{2mm}{\multirow{5}{*}{\rotatebox[origin=c]{90}{Instruments}}}
& K-Means & 0.678 & 0.181 & 0.705 & 0.213 & 0.764 \\
& Spectral & 0.686 & 0.196 & 0.713 & 0.229 & 0.767  \\
&  Agglo. & 0.707 & 0.226 & 0.719 & 0.257 & 0.776  \\
& SCL &  \underline{0.728} &  \underline{0.436} &  \underline{0.765} &  \underline{0.451} &  \underline{0.849} \\
& CACTUS & \textbf{0.786} & \textbf{0.553} & \textbf{0.817} & \textbf{0.578} & \textbf{0.883} \\

\midrule
\parbox[t]{2mm}{\multirow{5}{*}{\rotatebox[origin=c]{90}{Office}}}
& K-Means & 0.731 & 0.267 & 0.748 & 0.332 & 0.808 \\
& Spectral & 0.735 & 0.275 & 0.752 & 0.340 & 0.809 \\
&  Agglo. & 0.748 & 0.324 & 0.760 & 0.383 & 0.829 \\
& SCL &  \underline{0.772} & \underline{0.445} & \underline{0.792} & \underline{0.500} & \underline{0.866} \\
& CACTUS & \textbf{0.821} & \textbf{0.562} & \textbf{0.842} & \textbf{0.626} & \textbf{0.902} \\
\bottomrule
    \end{tabular}
\end{minipage}
\hspace{0.5ex}
\begin{minipage}{0.42\linewidth}
\centering
\small
        \caption{Results on validation set using different architectures for entity set encoder. Proposed method (section \ref{sec:caee}) is indicated by *. Augmented triplet loss is used to train all models.}
    \label{tab:set_encoders}
    
\begin{tabular}{lp{2.3cm}Hx{5ex}Hx{5ex}x{5ex}
x{3.2ex}}
\toprule
 & Set encoder & NMI & AMI & RI & ARI & F1 \\

\midrule
\parbox[t]{2mm}{\multirow{4}{*}{\rotatebox[origin=c]{90}{Arts}}}
& NIA & 0.687 & 0.354 & 0.730 & 0.409 & 0.826 \\
& SIA (KV-mean) & 0.697 & \underline{0.398} & 0.744 & 0.450 & 0.840 \\
& SIA (first) & 0.715 & 0.396 & 0.756 & 0.461 & 0.841 \\
& SIA (hid-mean)* & \textbf{0.732} & \underline{0.398} & \underline{0.764} & \underline{0.467} & \underline{0.845} \\
& FIA & \underline{0.719} & \textbf{0.423} & \textbf{0.765} & \textbf{0.494} & \textbf{0.851} \\

\midrule
\parbox[t]{2mm}{\multirow{4}{*}{\rotatebox[origin=c]{90}{Office}}}
& NIA & 0.779 & 0.442 & 0.801 & 0.495 & 0.867 \\
& SIA (KV-mean) & \underline{0.798} & 0.470 & 0.813 & 0.526 & 0.875 \\
& SIA (first) & 0.793 & \underline{0.493} & \underline{0.817} & 0.552 & \underline{0.881} \\
& SIA (hid-mean) & 0.789 & \textbf{0.513} & \underline{0.817} & \textbf{0.568} & \textbf{0.885} \\
& FIA & \textbf{0.805} & \underline{0.493} & \textbf{0.820} & \underline{0.553} & 0.879 \\
\bottomrule
    \vspace{1ex}

    \end{tabular}
\centering
    \includegraphics[scale=0.55]{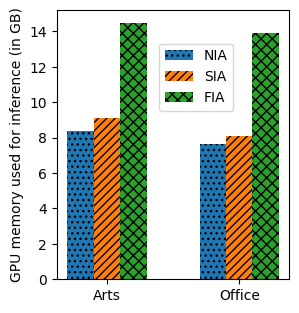}
    \captionof{figure}{GPU memory usage for inference using NIA, SIA (hid-mean), and FIA methods.}
    \label{fig:memory}
\end{minipage}
\end{table}

\subsection{Ablation studies}
We conduct ablation experiments to assess the effectiveness of the various proposed components, including context-aware entity embeddings, augmented triplet loss function, and self-supervised clustering task. For faster training, we utilize 3K training entity sets instead of the whole dataset for ablation studies. We focus on AMI, ARI, and F1 scores and skip NMI and RI as the latter can sometimes be high for random clusterings and are not adjusted for chance unlike AMI and ARI \citep{vinh2009information}.

\textbf{Set encoder } We compare five different methods of obtaining entity embeddings; the results are shown in Table \ref{tab:set_encoders}. The NIA, SIA (hid-mean), and FIA methods are described in Section \ref{sec:caee}. We explore two more scalable attention mechanisms: SIA (KV-mean) where keys and values are pooled instead of the hidden representations, and SIA (first) where the first token in each entity is used as the representative token for inter-entity attention. 
Both SIA and FIA methods obtain better results than NIA which demonstrates the importance of capturing the context given by an entity set. 
The FIA method achieves the best results on the Arts dataset, while SIA (hid-mean) achieves the best results on the Office dataset. 
Among the three SIA methods, SIA (hid-mean) yields the highest metrics on both datasets. 
Figure \ref{fig:memory} shows increasing GPU memory usage during inference from NIA to SIA (hid-mean) to FIA embedding methods. SIA achieves better results than FIA on the Office dataset, despite consuming 42\% less memory.

\begin{table}
\centering
\small
\makebox[0pt][c]{\parbox{1\textwidth}{%
    \begin{minipage}[b]{0.53\hsize}\centering
    \caption{Results on validation set using different supervised clustering loss functions for training. SIA (first) architecture is used for the set encoder.}
    \label{tab:losses}
    
\begin{tabular}{llHcHcc}
\toprule
 & Loss & NMI & AMI & RI & ARI & F1 \\

\midrule
\parbox[t]{2mm}{\multirow{4}{*}{\rotatebox[origin=c]{90}{Arts}}}
& cross-entropy & \textbf{0.725} & 0.374 & \textbf{0.756} & 0.441 & 0.832 \\
& structural loss & \underline{0.717} & 0.385 & 0.749 & 0.441 & 0.835 \\
& triplet & 0.710 & \underline{0.389} & 0.745 & \underline{0.444} & \underline{0.837} \\
& augmented triplet & 0.715 & \textbf{0.396} & \textbf{0.756} & \textbf{0.461} & \textbf{0.841} \\

\midrule
\parbox[t]{2mm}{\multirow{4}{*}{\rotatebox[origin=c]{90}{Office}}}
& cross-entropy & \underline{0.793} & 0.488 & \underline{0.818} & 0.548 & 0.876 \\
& structural loss & \textbf{0.801} & \underline{0.494} & \textbf{0.819} & \underline{0.549} & \textbf{0.881} \\
& triplet & 0.789 & \textbf{0.497} & 0.809 & 0.543 & 0.880 \\
& augmented triplet & \underline{0.793} & 0.493 & 0.817 & \textbf{0.552} & \textbf{0.881} \\

\bottomrule
\end{tabular}
    \end{minipage}
\hspace{1.5ex}
    \begin{minipage}[b]{0.43\hsize}\centering
    \caption{Results on validation set with and without self-supervision. SIA (hid-mean) architecture is used for the set encoder. (SS: Self-supervision)}
    \small
    \label{tab:pretraining}
    
    \begin{tabular}{lcHcHcc}
    \toprule
     & SS & NMI & AMI & RI & ARI & F1\\
    
\midrule
\parbox[t]{2mm}{\multirow{2}{*}{\rotatebox[origin=c]{90}{Arts}}}
& \xmark & \textbf{0.732} & 0.398 & \textbf{0.764} & 0.467 & 0.845 \\
& \cmark & 0.720 & \textbf{0.446} & 0.762 & \textbf{0.502} & \textbf{0.855} \\

\midrule
\parbox[t]{2mm}{\multirow{2}{*}{\rotatebox[origin=c]{90}{Office}}}
& \xmark & 0.789 & 0.513 & 0.817 & 0.568 & 0.885 \\
& \cmark & \textbf{0.811} & \textbf{0.552} & \textbf{0.834} & \textbf{0.608} & \textbf{0.894} \\
\noalign{\smallskip}
    \bottomrule
    \end{tabular}
    \end{minipage}
}}
\end{table}

\textbf{Loss function }
We compare different loss functions including the triplet and augmented triplet loss functions described in Section \ref{sec:loss}, the structural loss \cite{haponchyk2021supervised}, and binary cross-entropy loss for pairwise classification. The results are shown in Table \ref{tab:losses}. 
The augmented triplet loss achieves the highest AMI, ARI, and F1 scores on the Arts dataset and the highest ARI and F1 scores on the Office dataset.

\textbf{Self-supervision }
Table \ref{tab:pretraining} shows the clustering performance of our model with and without the proposed self-supervised pretraining phase as described in Section \ref{sec:pretraining}. 
We initialize the model with pretrained FlanT5 weights in both cases but include an extra dataset-specific pretraining phase before finetuning while using self-supervision. Self-supervised clustering improves AMI, ARI, and F1 on both datasets.

\vspace{-3pt}
\subsection{Qualitative analysis}
\begin{figure}
    \centering
    \includegraphics[scale=0.85]{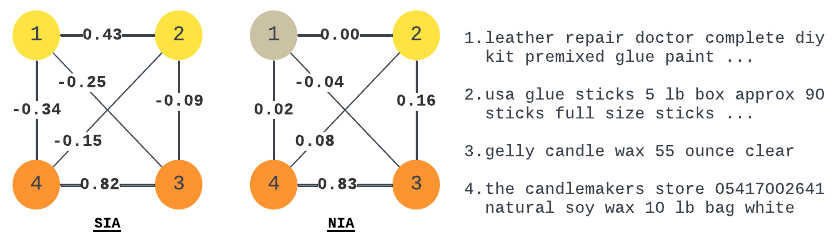}
    \caption{Case Study: Predicted clusterings with pairwise similarities using SIA and NIA methods. The SIA method correctly identifies the common cluster membership of the first two entities where NIA fails. The stopping threshold for agglomerative clustering is chosen based on the results of the validation set.}
    \label{fig:case}
\end{figure}

\vspace{-4pt}
We will qualitatively demonstrate the significance of context-aware embeddings using an example. 
Referring to Figure \ref{fig:case}, using SIA embeddings, our model accurately identifies two products each under the `Glue Products' and `Candle Making Supplies' clusters. 
However, with NIA embeddings, the model fails to capture the similarity between the two glue products. 
Specifically, in the NIA embeddings, the first product, a leather repair glue paint, is placed closer to other leather repair products in the universal entity set but far away from products containing glue sticks. 
The SIA approach leverages the context provided by the current entity set and places the leather repair glue paint and glue sticks (the first two entities) in the same cluster.

\vspace{-1ex}
\section{Conclusion}
\vspace{-1ex}
This paper presented a novel approach for supervised clustering of entity subsets using context-aware entity embeddings from LLMs. 
Context-awareness is achieved through a scalable inter-entity attention mechanism that facilitates interactions among different entities at each layer of the LLM. 
We also proposed an augmented triplet loss to address challenges encountered when directly applying triplet loss to supervised clustering. 
A self-supervised clustering task is introduced by drawing inspiration from text-augmentation techniques, which helps in enhancing the fine-tuning performance. 
We demonstrated that by integrating the proposed components, our model outperforms existing methods by a significant margin on extrinsic clustering evaluation metrics. 
Future research could investigate alternative techniques for inter-entity attention, explore additional loss functions and self-supervision tasks, and extend the current work to more application domains.



\bibliography{colm2024_conference}
\bibliographystyle{colm2024_conference}

\newpage
\appendix

\appendix

\setcounter{table}{0}
\setcounter{figure}{0}
\renewcommand*\thetable{A\arabic{table}}
\renewcommand*\thefigure{B\arabic{figure}}
\section*{\fontsize{12}{15}\selectfont APPENDIX}

\section{Collecting ground truth for supervised clustering} \label{sec:ground}
We obtain ground truth clusterings for entity subsets by prompting a proprietary closed-source LLM ($\text{LLM}_\text{c}$). 
Given an entity set $E=\{e_1,...,e_{|E|}\}$, we use the following prompt as input to $\text{LLM}_\text{c}$.
\begin{quote}
Cluster these products:\\
\texttt{text$(e_1)$}\\
\vdots\\
\texttt{text$(e_{|E|})$} \\
For each cluster, the answer should contain a meaningful cluster title and the products in that cluster. Do not provide any explanation.
\end{quote}
Based on observing the text outputs for a few cases, we implement a parsing algorithm to convert these outputs into clusterings. 
We discard outputs that are either empty or cannot be parsed; these constitute less than 0.08\% of entity sets. 
Additionally, entities that do not appear in the parsed clusterings (less than 3\% per sequence on average) are removed. 

\section{Self-supervised clustering} \label{sec:pretraining}
To improve the generalization ability of our model ($\text{LLM}_\text{o}$), especially when a limited number of entity sets with ground truth clusterings are available, we introduce a self-supervised clustering task that artificially creates ground truth clusterings using text augmentation ideas from contrastive learning. To create a training sample for self-supervision, we first randomly sample the number of clusters and the size of each cluster, then randomly sample a seed entity (from the universal set of entities) for each cluster, and populate each cluster with different transformations of the seed entity. The transformations are obtained by randomly dropping words from the original description of the seed entity. The same augmented triplet loss function used for supervised clustering is used for self-supervised pretraining as well. The number of clusters is sampled from $\mathcal{U}(2,10)$, cluster sizes are sampled from $\mathcal{U}(1,5)$, and the fraction of words to drop from a seed entity is sampled from $\mathcal{U}(0.2,0.7)$. We run self-supervised pretraining separately for each dataset using the universal set of entities from the corresponding dataset.

\section{Implementation details}
We implemented the proposed methods and baselines in python using the HuggingFace Transformers library \citep{huggingface}. We adapted the T5ForConditionalGeneration class to implement the Scalable Inter-entity Attention (SIA) method. Our experiments were run on an Ubuntu 20.04.6 LTS server using a single NVIDIA Quadro RTX GPU. For both pretraining and finetuning, we used a learning rate of 1e-4 and a batch size of 4 during training and a batch size of 16 for inference. The finetuning was run for 10 epochs, with an evaluation on the validation set performed after each epoch. The checkpoint at the epoch with the best combined (sum) NMI, AMI, RI, and ARI on the validation set is selected for evaluation on the test set. 
For all supervised methods, we run average-link agglomerative clustering using predicted pairwise similarities, varying the threshold from -1 to 1 in increments of 0.1. The optimal threshold was selected based on performance on the validation set. 
The pretraining was run for 20K training batches. To ensure reproducibility, we seeded all random number generators before each experiment.

For SCL baseline, we used the hyperparameter values of C and r as 0.15 and 0.5, respectively, as recommended in \citep{haponchyk2021supervised}. 
For both triplet loss and augmented triplet loss, we set the margin to 0.3, initializing the neutral similarity $s_{neut}$ to 0 for augmented triplet loss. 
For unsupervised clustering methods, we computed entity embeddings similarly to NIA but using the pretraining Flan-T5-base weights. We used implementations of KMeans, Spectral, and Agglomerative clustering algorithms from scikit-learn \citep{scikit-learn}. 
The code for this work is provided at 
\url{https://github.com/amazon-science/context-aware-llm-clustering}.


\section{Metrics}

\begin{wrapfigure}{r}{0.5\textwidth}
\vspace{-4ex}
\footnotesize
\begin{tabular}{c|c|c|c|c|c}
    & \multicolumn{4}{c|}{Clustering B} \\
    \hline
    \multirow{4}{*}{Clustering A}
    & $n_{11}$ & $n_{12}$ & $\hdots$ & $n_{1s}$ &$a_1$ \\
    \cline{2-6}
    & $n_{21}$ & $n_{22}$ & $\hdots$ & $n_{2s}$ &$a_2$ \\
    \cline{2-6}
    & $\vdots$ &$\vdots$ &$\ddots$ &$\vdots$ &$\vdots$ \\
    \cline{2-6}
    & $n_{r1}$ & $n_{r2}$ & $\hdots$ & $n_{rs}$ &$a_r$ \\
    \hline
    & $b_1$ & $b_2$ & $\hdots$ &$b_s$ & $n$\\
    \end{tabular}
\end{wrapfigure}
Consider a ground truth clustering `A' and a predicted clustering `B' with the contingency matrix shown on the right. 
The external clustering metrics used in the paper are defined as follows. \\ \\ \\ \\

\vspace{-8ex}
\textbf{Normalized Mutual Information}: $NMI=\frac{2\,MI(A,B)}{H(A)+H(B)} \in [0,1]$, where $H(A)=-\sum_{i=1}^r\frac{a_i}{n}log\big(\frac{a_i}{n}\big)$ and $H(B)=-\sum_{j=1}^s\frac{b_j}{n}log\big(\frac{b_i}{n}\big)$ are the entropies of clusterings `A' and `B', and $MI(A,B)=\sum_{i=1}^r\sum_{j=1}^s \frac{n_{ij}}{n} log\big(\frac{n_{ij}n}{a_ib_j}\big) $ is the mutual information between the two clusterings.

\textbf{Adjusted Mutual Information:} $AMI=\frac{MI-\mathbb{E}[MI(A,B)]}{\frac{H(A)+H(B)}{2}-\mathbb{E}[MI(A,B)]}$ adjusts MI for chance. Assuming the generalized hypergeometric distribution, the expected value of MI is given by $\mathbb{E}[MI(A,B)]=\sum_{i=1}^r\sum_{j=1}^s \sum_{n_{ij}=max(a_i+b_j-n,0)}^{min(a_i,b_j)} \frac{n_{ij}}{n} log\big(\frac{nn_{ij}}{a_ib_j}\big)\frac{a_i!b_j!(n-a_i)!(n-b_j)!}{n!n_{ij}!(a_i-n_{ij})!(b_j-n_{ij})!(n-a_i-b_j+n_{ij})!}$
\citep{vinh2009information}.

\textbf{Rand Index}: $RI = \frac{\alpha +\beta}{{n\choose 2}} \in [0,1]$ where $\alpha$ is the number of pairs within same cluster in both A and B, and $\beta$ is the number of pairs in different clusters in both A and B. 

\textbf{Adjusted Rand Index:}
$ARI=\frac{RI-\mathbb{E}[RI]}{max[RI] - \mathbb{E}[RI]} \leq 1$ adjusts RI for chance. Assuming the generalized hypergeometric distribution, we can show that $ARI = \frac{\sum_{ij}{n_{ij}\choose 2} - \big[\sum_i {a_i\choose 2}\sum_j{b_j\choose 2}\big]/{n\choose 2}}
{\frac{1}{2}\big[ \sum_i {a_i\choose 2} + \sum_j{b_j\choose 2} \big] -\big[\sum_i {a_i\choose 2}\sum_j{b_j\choose 2}\big]/{n\choose 2}}$ \citep{hubert1985comparing}.

\textbf{F1-score: } This is calculated as the harmonic mean of precision and recall, where $precision=\frac{\sum_j max(\{n_{1j},...,n_{rj}\})}{n}$ and $recall=\frac{\sum_i max(\{n_{i1},...,n_{is}\})}{n}$\citep{haponchyk2018supervised}.

\begin{table}[]
    \centering
    \caption{Cross-dataset evaluation: Models are trained on `source' dataset and evaluated on the test set of `target' dataset. (*Similar to Table 2 in the main paper, for the proprietary Gifts dataset, we report improvements against K-Means.)} \label{tab:cross}
    \footnotesize
    \begin{tabular}{|c|c|HcHcc|HcHcc|}
    \hline
    Source & Target & \multicolumn{5}{c|}{SCL} &  \multicolumn{5}{c|}{CACTUS} \\
    \hline 
    & & NMI & AMI & RI & ARI & F1 & NMI & AMI & RI & ARI & F1  \\
    \hline
\multirow{4}{*}{Gifts}
& Arts & \textbf{0.680} & 0.270 & 0.711 & 0.340 & \textbf{0.81} & 0.677 & \textbf{0.304} & \textbf{0.712} & \textbf{0.361} & \textbf{0.81}\\
& Games & 0.639 & 0.250 & 0.686 & 0.277 & \textbf{0.795} & \textbf{0.673} & \textbf{0.279} & \textbf{0.712} & \textbf{0.295} & 0.793\\
& Instruments & \textbf{0.668} & 0.244 & \textbf{0.695} & 0.273 & 0.791 & 0.633 & \textbf{0.292} & 0.685 & \textbf{0.297} & \textbf{0.797}\\
& Office & 0.714 & 0.354 & 0.744 & 0.414 & 0.843 & \textbf{0.74} & \textbf{0.401} & \textbf{0.763} & \textbf{0.447} & \textbf{0.848}\\
\hline
\multirow{4}{*}{Arts}
& Gifts* & 0.026 & 0.030 & -0.046 & 0.035 & 0.052 & \textbf{0.082} & \textbf{0.103} & \textbf{0.016} & \textbf{0.101} & \textbf{0.072}\\
& Games & \textbf{0.704} & 0.245 & 0.724 & 0.274 & 0.789 & 0.689 & \textbf{0.284} & \textbf{0.727} & \textbf{0.313} & \textbf{0.802}\\
& Instruments & 0.7 & 0.240 & 0.716 & 0.270 & 0.784 & \textbf{0.704} & \textbf{0.331} & \textbf{0.735} & \textbf{0.354} & \textbf{0.812}\\
& Office & 0.76 & 0.355 & 0.769 & 0.413 & 0.836 & \textbf{0.772} & \textbf{0.416} & \textbf{0.791} & \textbf{0.478} & \textbf{0.859}\\
\hline
\multirow{4}{*}{Games}
& Gifts* & 0.01 & 0.016 & -0.058 & 0.025 & \textbf{0.051} & \textbf{0.068} & \textbf{0.064} & \textbf{0.003} & \textbf{0.063} & 0.046\\
& Arts & \textbf{0.684} & 0.274 & 0.713 & 0.337 & \textbf{0.805} & 0.683 & \textbf{0.276} & \textbf{0.715} & \textbf{0.341} & 0.802\\
& Instruments & \textbf{0.698} & 0.226 & 0.713 & 0.257 & 0.780 & 0.695 & \textbf{0.274} & \textbf{0.723} & \textbf{0.304} & \textbf{0.797}\\
& Office & \textbf{0.749} & 0.347 & \textbf{0.764} & 0.406 & \textbf{0.838} & 0.725 & \textbf{0.351} & 0.756 & \textbf{0.407} & 0.837\\
\hline
\multirow{4}{*}{Instruments}
& Gifts* & -0.001 & 0.014 & -0.067 & 0.022 & 0.052 & \textbf{0.059} & \textbf{0.087} & \textbf{-0.002} & \textbf{0.086} & \textbf{0.069}\\
& Arts & 0.696 & 0.267 & 0.716 & 0.333 & 0.797 & \textbf{0.698} & \textbf{0.282} & \textbf{0.72} & \textbf{0.350} & \textbf{0.804}\\
& Games & \textbf{0.703} & 0.251 & 0.723 & 0.280 & 0.790 & 0.687 & \textbf{0.291} & \textbf{0.726} & \textbf{0.321} & \textbf{0.804}\\
& Office & 0.752 & 0.374 & 0.768 & 0.429 & 0.844 & \textbf{0.763} & \textbf{0.398} & \textbf{0.782} & \textbf{0.457} & \textbf{0.853}\\
\hline
\multirow{4}{*}{Office}
& Gifts* & 0.021 & 0.025 & -0.051 & 0.032 & 0.055 & \textbf{0.045} & \textbf{0.069} & \textbf{-0.018} & \textbf{0.073} & \textbf{0.066}\\
& Arts & 0.676 & 0.292 & 0.710 & 0.353 & 0.810 & \textbf{0.701} & \textbf{0.322} & \textbf{0.732} & \textbf{0.390} & \textbf{0.818}\\
& Games & 0.61 & 0.226 & 0.670 & 0.248 & 0.788 & \textbf{0.686} & \textbf{0.276} & \textbf{0.723} & \textbf{0.306} & \textbf{0.802}\\
& Instruments & \textbf{0.715} & 0.167 & 0.716 & 0.210 & 0.756 & 0.698 & \textbf{0.295} & \textbf{0.728} & \textbf{0.325} & \textbf{0.804}\\
\hline

    \end{tabular}
    \label{tab:my_label}
\end{table}

\section{Cross-dataset evaluation}
Table \ref{tab:cross} shows the performance of SCL\citep{Barnabo2023} and CACTUS (ours) when trained on one dataset and evaluated on another. CACTUS outperforms SCL in most cases. Moreover, CACTUS achieves higher AMI, ARI, and F1 compared to the unsupervised baselines (Table 2 in main paper) across all datasets, except for the F1 score when trained on Gifts and evaluated on Games. On the other hand, SCL fails to outperform unsupervised baselines when the source dataset is Games, and when the source dataset is Instruments and the target is Office. This shows that context-awareness enables our method to learn to identify more general clustering patterns in input entity subsets.

\end{document}